\documentclass[10pt,twocolumn,letterpaper]{article}

\usepackage{cvpr}              

\definecolor{cvprblue}{rgb}{0.21,0.49,0.74}
\usepackage[pagebackref,breaklinks,colorlinks,allcolors=cvprblue]{hyperref}
\usepackage[accsupp]{axessibility} 
\usepackage{multirow}
\usepackage{pifont}
\usepackage{colortbl}
\usepackage{subcaption}
\usepackage{tabularx}
\def\confName{CVPR}
\def\confYear{2025}

\title{OSGNet @ Ego4D Episodic Memory Challenge 2025 
}

\author{
Yisen Feng$^{1}$, Haoyu Zhang$^{1\,2}$, Qiaohui Chu$^{1\,2}$, Meng Liu$^{3}$, Weili Guan$^{1}$, \\ Yaowei Wang$^{1\,2}$, Liqiang Nie$^{1}$\\
$^1$Harbin Institute of Technology (Shenzhen) \qquad  $^2$Pengcheng Laboratory    \\$^3$Shandong Jianzhu University\\
{\tt\small \{yisenfeng.hit, zhang.hy.2019, qiaohuichu8599, mengliu.sdu, honeyguan, nieliqiang\}@gmail.com;} \\ {\tt\small wangyw@pcl.ac.cn}
}

\begin{document}
\maketitle
\begin{abstract}
In this report, we present our champion solutions for the three egocentric video localization tracks of the Ego4D Episodic Memory Challenge at CVPR 2025. All tracks require precise localization of the interval within an untrimmed egocentric video. Previous unified video localization approaches often rely on late fusion strategies, which tend to yield suboptimal results. To address this, we adopt an early fusion-based video localization model to tackle all three tasks, aiming to enhance localization accuracy. Ultimately, our method achieved first place in the Natural Language Queries, Goal Step, and Moment Queries tracks, demonstrating its effectiveness.
Our code can be found at \url{https://github.com/Yisen-Feng/OSGNet}.

\end{abstract}    
\section{Introduction}
\label{sec:intro}
Egocentric video moment localization~\cite{feng2024objectnlq,hou2023groundnlq}, as the basis of many egocentric video understanding tasks, underpins a range of innovative applications~\cite{10.1145/3474085.3475234,10239469,guan2022bi,zhang2021multimodal,zhang2023uncovering,chu2025technicalreportego4dlongterm}, such as intelligent assistant systems~\cite{zhang2024hcqa,zhang2025hcqa} in smart glasses and memory retrieval modules for embodied AI~\cite{wang2025time,zhang2025exo2ego}. However, unlike traditional exocentric video moment localization~\cite{liu2018attentive,liu2018cross}, this task faces unique challenges due to the rapid viewpoint shifts and the emphasis on fine-grained objects~\cite{feng2025object}.

In the Episodic Memory Challenge of Ego4D~\cite{grauman2022ego4d},  egocentric video moment localization is framed in three task variants. 1) Natural Language Queries (NLQ), involves locating the relevant moment in an egocentric video given a natural language query. 2) GoalStep-Step Grounding~\cite{song2024ego4d}, requires identifying the egocentric video moment corresponding to a described step. 3) Moment Queries (MQ),  focuses on localizing predefined actions within the egocentric video. Although these tasks differ in input formulation, they all share the common objective of precisely identifying temporal intervals within egocentric videos~\cite{pmlr-v235-zhang24aj} and often rely on similar model architectures.

Recent works~\cite{zeng2025unimd,zhang2025timeloc} explore unified frameworks capable of handling different localization tasks simultaneously. However, in pursuit of efficiency, these approaches typically adopt late fusion for multimodal fusion, resulting in suboptimal localization performance. To address this limitation, we adapt our previously proposed state-of-the-art early fusion methods~\cite{feng2025object} for egocentric moment localization, aiming to achieve better localization performance.

Based on our powerful model, we surpassed last year's champion solution in all three tracks and won the first place in this year's challenge. Specifically, on NLQ, our model outperforms EgoVideo~\cite{pei2024egovideo}, achieving a 2.47\% improvement in Rank@1 at IoU=0.5. On Goal Step, our model surpasses BayesianVSLNet~\cite{plou2024carlor} with a 6.84\% increase in Rank@1 at IoU=0.3. On MQ, our model achieves a 1.79\% improvement in mean Average Precision (mAP) over CausalTAD~\cite{liu2024harnessing}.

\section{Methodology}

\subsection{Task Definition}
The video moment localization task can be formally defined as follows: Given a video $\mathcal{V} = \{v_1, v_2, \cdots, v_N\}$ consisting of $N$ frames and a natural language query (or step description, action category, etc.) $\mathcal{Q} = \{w_1, w_2, \cdots, w_L\}$ containing $L$ words, the objective is to identify the temporal segment $[t_s, t_e]$ within the video that best corresponds to the semantic content of the query.


\begin{figure*}[t]
  \centering
  \includegraphics[width=\linewidth]{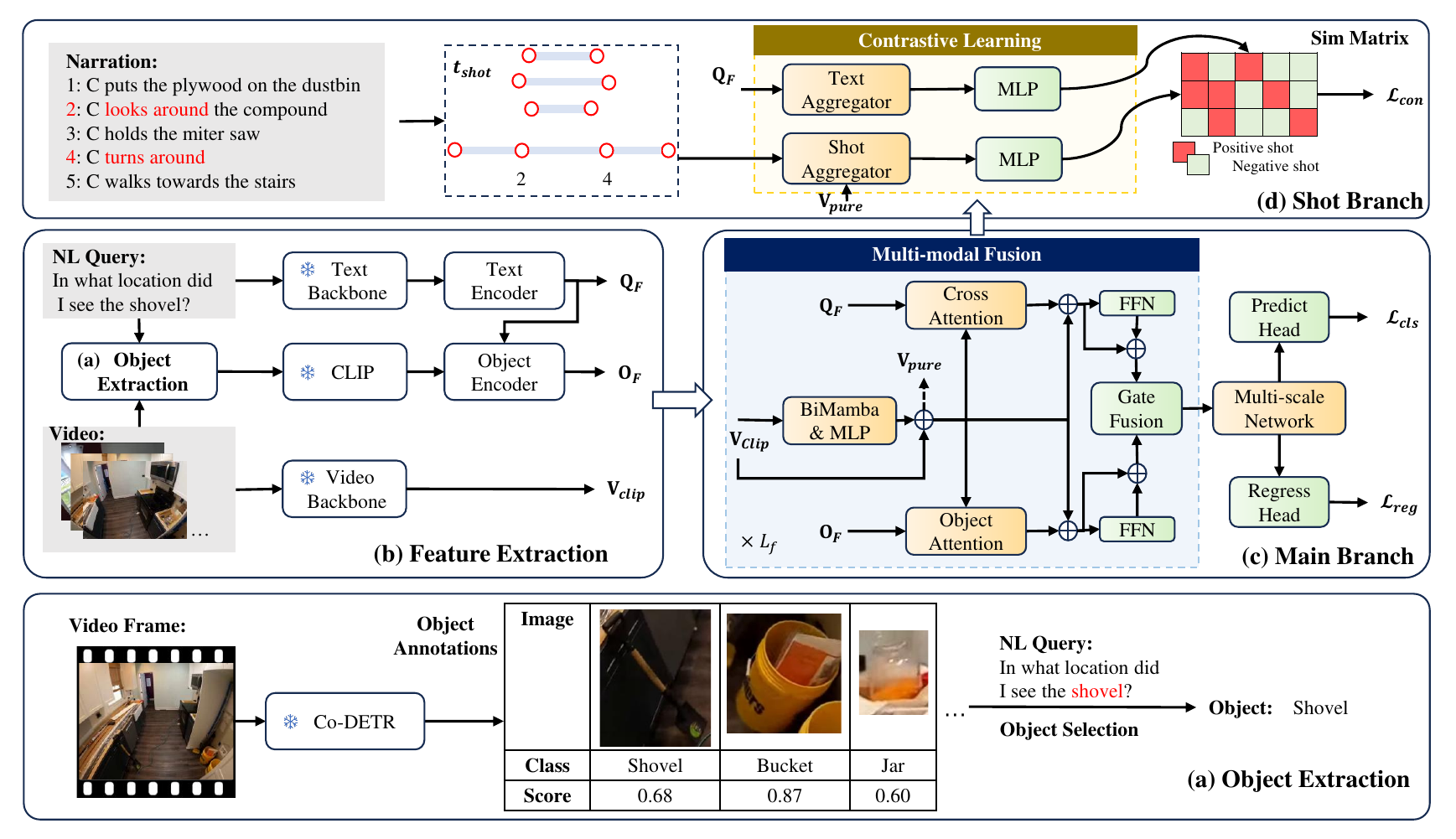}
  \vspace{-4ex}
 \caption{The framework of OSGNet~\cite{feng2025object}.}

  \vspace{-2ex}
  \label{fig: model}
\end{figure*}
\subsection{Track 1: Natural Language Queries}
\textbf{Approach.}
As illustrated in Figure~\ref{fig: model}, we adopt our previously proposed grounding model, OSGNet~\cite{feng2025object}, to address the challenge of insufficient fine-grained object information in the NLQ task. OSGNet enhances video representations by incorporating object-level details through the extraction and encoding of features from video, text, and detected objects, thereby enabling more accurate moment localization. The framework consists of two key components: a main branch that fuses video, text, and object features to improve localization precision, and a shot branch that processes video-only features to generate shot-level representations, leveraging contrastive learning with textual inputs to strengthen cross-modal alignment. For further architectural specifics, please refer to our original work~\cite{feng2025object}.


\noindent \textbf{Implementation Details.}
1) \textit{Feature Extraction:} We utilize EgoVideo~\cite{pei2024egovideo} and InternVideo~\cite{chen2022internvideo} to extract video features by dividing each video into multiple non-overlapping snippets, each consisting of 16 frames. 
For text feature extraction, we use EgoVideo and CLIP independently, representing each word in the sentence with its corresponding feature and discarding the additional [CLS] token.
2) \textit{Training Setup:} Following the experimental setup of OSGNet, we first train OSGNet without the shot branch and object features (named OSGNet-baseline) utilizing the NaQ~\cite{ramakrishnan2023naq} pretraining strategy to improve efficiency. In the pretraining phase, learning rate is set to 8e-4 with a batch size of 16. During the main training phase, the learning rate is increased to 1.6e-3 while maintaining the same batch size. Our model is trained for 10 epochs in total, with the first 4 epochs designated as a warm-up phase.

\noindent\textbf{Results.}
\begin{table*}[t]
\caption{Performance on NLQ. $^\dagger$ results utilize the ensemble strategy.}

  \vspace{-2ex}
  \label{tab: NLQ}
  \centering
  \begin{tabular}{lcccccccc}
    \toprule
    \multirow{3}{*}{Method}&\multicolumn{4}{c}{Validation}  & \multicolumn{4}{c}{Test}        \\
    
    & \multicolumn{2}{c}{R@1}&\multicolumn{2}{c}{R@5}&\multicolumn{2}{c}{R@1}&\multicolumn{2}{c}{R@5}\\
       & 0.3 & 0.5 & 0.3 & 0.5 &0.3 & 0.5 & 0.3 & 0.5 \\
    \midrule
   
    EgoVideo~\cite{pei2024egovideo}&28.65& 19.73& 53.30& 40.42& 25.07& 17.31& 40.88& 29.67 \\	
    EgoVideo$^\dagger$~\cite{pei2024egovideo}&-& -& -& -& 28.05& 19.31& 44.16 &31.37 \\	 
     OSGNet~\cite{feng2025object}& 32.56   &  22.74   &  59.82   &  46.35 &28.50&20.05&46.00&32.97\\
     \rowcolor{gray!15}OSGNet$^\dagger$~\cite{feng2025object}&- &- &- 	&- &\textbf{30.19} &\textbf{21.78} &\textbf{47.50} &\textbf{35.84}  \\
    \bottomrule
  \end{tabular}
\end{table*}
As shown in Table~\ref{tab: NLQ}, our model outperforms last year’s champion solution using only the training set and a single model. Subsequently, we further improved performance by integrating multiple model variants and incorporating the validation set into the training process, achieving a 1.73\% gain in R@1, IoU=0.5 over the single model. Specifically, the integrated ensemble comprises four model variants: (1) the original OSGNet, (2) OSGNet without shot branches, (3) OSGNet utilizing InternVideo features, and (4) OSGNet augmented with additional cross-modal attention layers in the multi-scale network.

\noindent\textbf{Case Analysis.}
\begin{figure}[t]
  \centering
  \includegraphics[width=\linewidth]{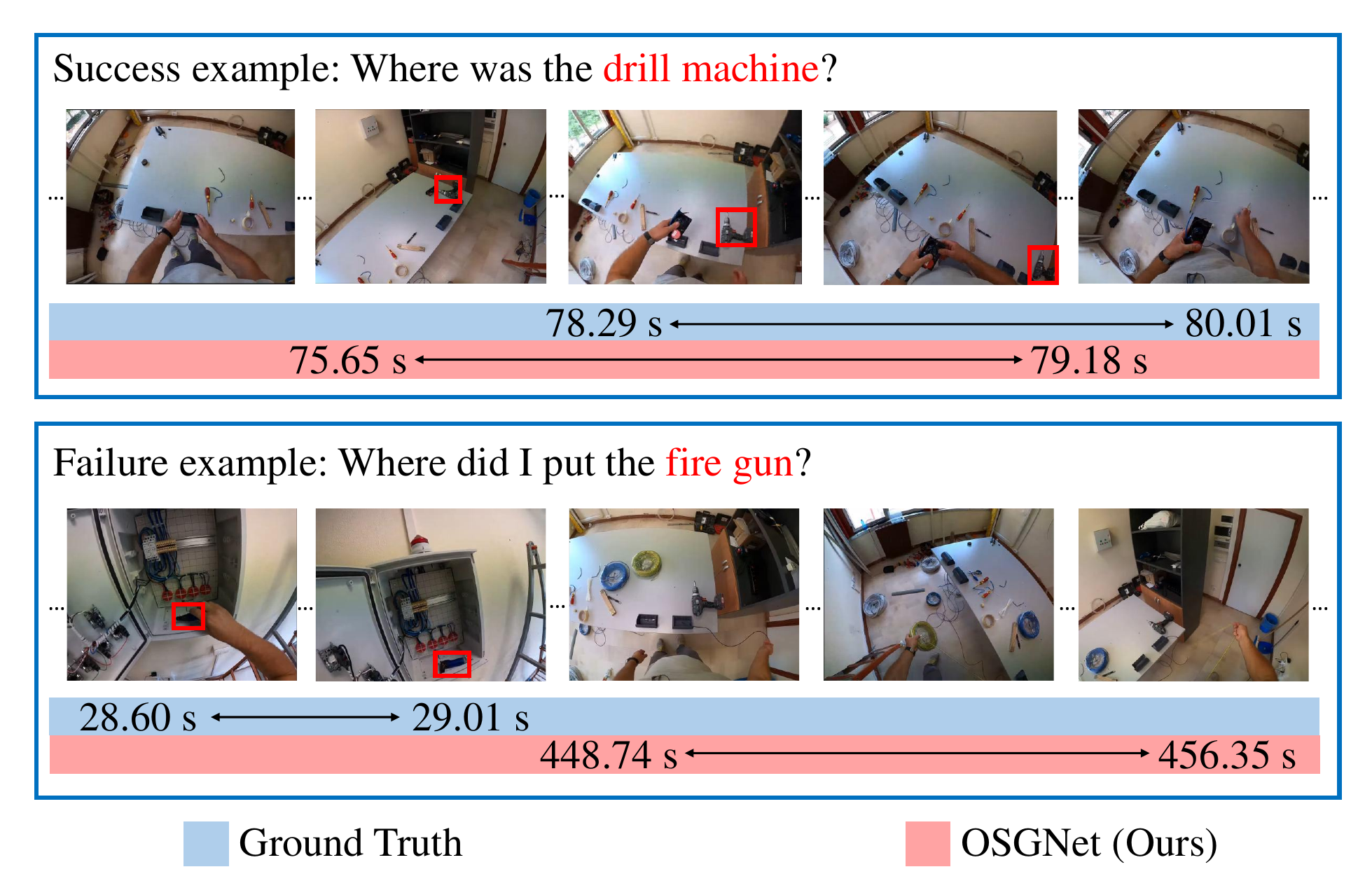}
  \vspace{-4ex}
  \caption{Two examples on the validation set of NLQ.}
  \vspace{-2ex}
  \label{fig: NLQ}
\end{figure}
As shown in Figure~\ref{fig: NLQ}, in the first example, our model accurately identifies the time interval in which the drill machine appears, demonstrating its effective use of fine-grained object information. However, in the second example, the model fails to correctly localize the fire gun, as the object detector has not been trained on this object category. This highlights our method's limitation of relying on the coverage and capability of the underlying object detector.

\subsection{Track 2: GoalStep - Step Grounding}
\textbf{Approach.}
Similar to NLQ, we utilize OSGNet-baseline for grounding, and employ EgoVideo for extracting video and text features.

\noindent\textbf{Implementation Details.}
We follow the same feature extraction and pretraining settings as used in NLQ. During the main training phase, we increase the learning rate to 1.6e-3 and use a batch size of 32. Given the excessive length of input videos, we apply random cropping and cap the number of video frame features at 2,560 to manage computational cost. During inference, we incorporate the order prior as introduced in BayesianVSLNet~\cite{plou2024carlor}.


\noindent\textbf{Results.}
\begin{table}[t]
\caption{R@1 Performance on Goal Step. $^\dagger$ results utilize the ensemble strategy.}

  \vspace{-2ex}
  \label{tab: GoalStep}
  \centering
  \resizebox{0.45\textwidth}{!}{
  \begin{tabular}{lcccc}
    \toprule
    \multirow{2}{*}{Method}  & \multicolumn{2}{c}{Validation}  & \multicolumn{2}{c}{Test}        \\

       & 0.3 & 0.5 & 0.3 & 0.5 \\
    \midrule

    BayesianVSLNet~\cite{plou2024carlor}&18.15&8.97&35.18&20.48\\
     OSGNet~\cite{feng2025object}& 44.27&	37.37&	40.51&	31.55\\
     \rowcolor{gray!15}OSGNet$^\dagger$~\cite{feng2025object} &- &-&\textbf{42.02} &\textbf{32.83}  \\
    \bottomrule
  \end{tabular}
  }
\end{table}
Table~\ref{tab: GoalStep} presents the performance comparison on the Goal Step. Our single model outperforms last year’s champion solution by a substantial margin of 5.33\% in R@1, IoU=0.3. The performance is further enhanced by an additional 1.51\% through model ensemble and the incorporation of the validation set. 
Specifically, the ensemble combines three model variants: the OSGNet-baseline, the OSGNet-baseline with full-length video feature input, and the OSGNet-baseline augmented with additional cross-modal attention layers in the multi-scale network (denoted as OSGNet-baseline*).

\noindent\textbf{Case Analysis.}
\begin{figure}[t]
  \centering
  \includegraphics[width=\linewidth]{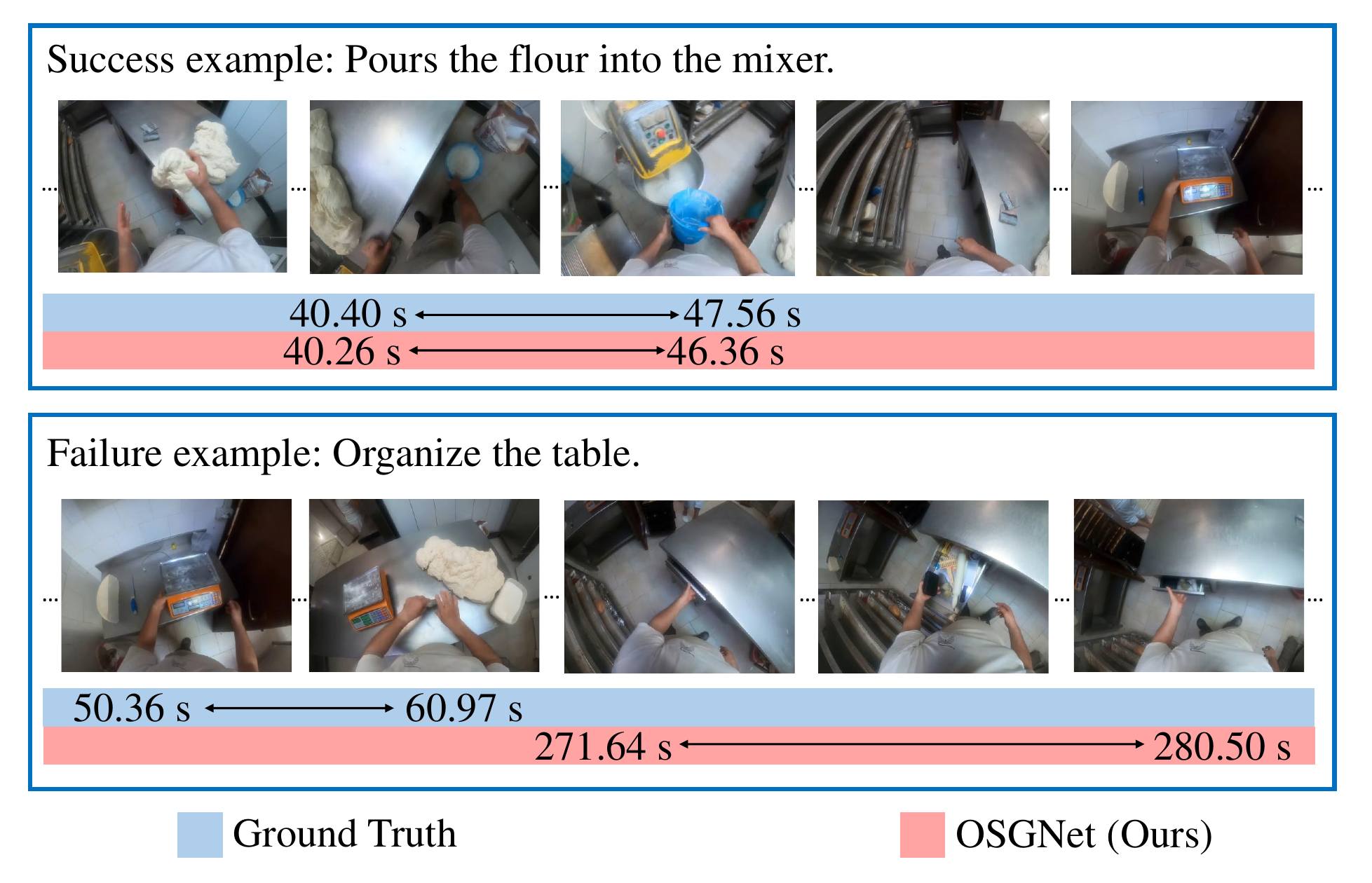}
  \vspace{-4ex}
  \caption{Two examples on the validation set of Goal Step.}
  \vspace{-2ex}
  \label{fig: GS}
\end{figure}
As shown in Figure~\ref{fig: GS}, in the first example, our model accurately localizes the action of pouring the flour into the mixer, demonstrating its strong localization capability. However, in the second example, the model fails to correctly identify the action of organizing the table. We speculate that while the model recognizes both the action ``organize'' and the object ``table'', it misinterprets the action of organizing the drawer as organizing the table, indicating a limited understanding of the subject-object relationship within the action.

\subsection{Track 3: Moment Queries}
\textbf{Approach.}
We employ the same model and features for grounding as GoalStep. Since MQ is a temporal action localization task, we convert it into a text-based video moment localization task by treating all predefined action categories as text.

\noindent\textbf{Implementation Details.}
We follow the same settings as NLQ to extract features and pretrain. During training, the learning rate is set to 4e-4 with a batch size of 32.

\noindent\textbf{Results.}
\begin{table}[t]
\caption{Performance on MQ. $^\dagger$ results utilize the ensemble strategy. R1@0.5 represents R@1, IoU=0.5 metric.}

  \vspace{-2ex}
  \label{tab: MQ}
  \centering
  \begin{tabular}{lcccc}
    \toprule
    \multirow{2}{*}{Method}  & \multicolumn{2}{c}{Validation}  & \multicolumn{2}{c}{Test}        \\

       & mAP& R1@0.5 & mAP& R1@0.5 \\
    \midrule

    CausalTAD&33.59&-&34.99$^\dagger$&52.83$^\dagger$\\
     OSGNet~\cite{feng2025object}& 35.92&	52.16&	35.31&	52.83\\
     \rowcolor{gray!15}OSGNet$^\dagger$~\cite{feng2025object} &- &-&\textbf{36.78} &\textbf{54.08}  \\
    \bottomrule
  \end{tabular}
\end{table}
Table~\ref{tab: MQ} presents the performance comparison on the MQ benchmark. Our single model, trained solely on the training set, outperforms last year’s winning solution. By incorporating the validation set and applying model ensembling, the performance is further improved by 1.47\% in mAP. Specifically, the ensemble includes three model variants: the OSGNet-baseline, the OSGNet-baseline trained only on the training set, and the OSGNet-baseline* variant, which includes additional cross-modal attention layers and is also trained solely on the training set.

\noindent\textbf{Case Analysis.}
\begin{figure}[t]
  \centering
  \includegraphics[width=\linewidth]{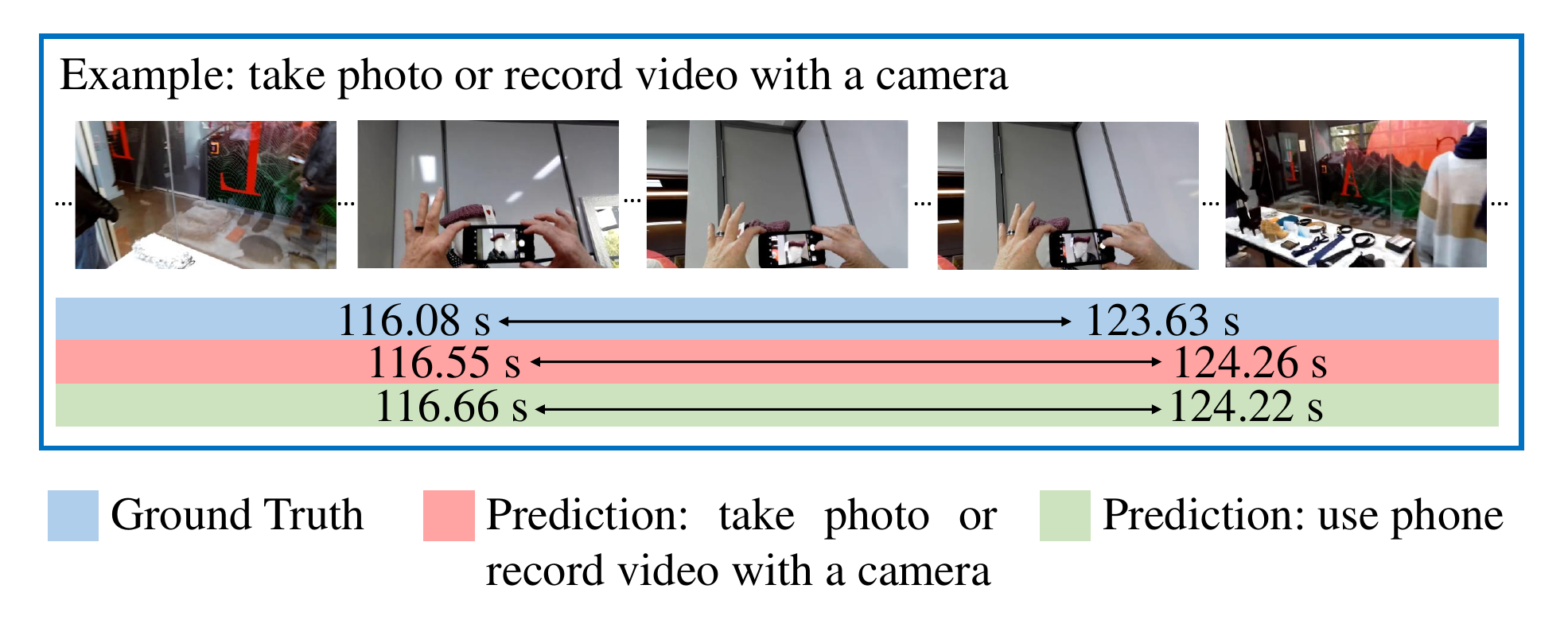}
  \vspace{-4ex}
  \caption{Example on the validation set of MQ.}
  \vspace{-2ex}
  \label{fig: MQ}
\end{figure}
As shown in Figure~\ref{fig: MQ}, our model demonstrates strong localization capabilities, accurately identifying the video moment corresponding to the correct label. However, it also assigns the wrong label to the same segments, which we suspect stems from an insufficient understanding of the video content.

\section{Conclusion}
This report presents our methods and results on the NLQ, Goal Step, and MQ tracks of the Ego4D Challenge at CVPR 2025. Our primary contribution is the application of a state-of-the-art retrieval-based model to temporal action localization tasks, which achieves superior performance compared to conventional localization methods. Despite its effectiveness, our approach has certain limitations. Reformulating action localization as a retrieval problem leads to increased training and inference time, suggesting that future work should explore strategies to better balance performance and computational efficiency.



{
    \small
    \bibliographystyle{ieeenat_fullname}
    \bibliography{main}

\begin{thebibliography}{25}
\providecommand{\natexlab}[1]{#1}
\providecommand{\url}[1]{\texttt{#1}}
\expandafter\ifx\csname urlstyle\endcsname\relax
  \providecommand{\doi}[1]{doi: #1}\else
  \providecommand{\doi}{doi: \begingroup \urlstyle{rm}\Url}\fi

\bibitem[Chen et~al.(2022)Chen, Xing, Chen, Wang, Li, Li, Liu, Wang, Zheng, Huang, et~al.]{chen2022internvideo}
Guo Chen, Sen Xing, Zhe Chen, Yi Wang, Kunchang Li, Yizhuo Li, Yi Liu, Jiahao Wang, Yin-Dong Zheng, Bingkun Huang, et~al.
\newblock Internvideo-ego4d: {{A}} pack of champion solutions to ego4d challenges.
\newblock \emph{arXiv preprint arXiv:2211.09529}, pages 1--11, 2022.

\bibitem[Chu et~al.(2025)Chu, Zhang, Feng, Liu, Guan, Wang, and Nie]{chu2025technicalreportego4dlongterm}
Qiaohui Chu, Haoyu Zhang, Yisen Feng, Meng Liu, Weili Guan, Yaowei Wang, and Liqiang Nie.
\newblock Technical report for ego4d long-term action anticipation challenge 2025.
\newblock \emph{arXiv preprint arXiv:2506.02550}, 2025.

\bibitem[Feng et~al.(2024)Feng, Zhang, Xie, Li, Liu, and Nie]{feng2024objectnlq}
Yisen Feng, Haoyu Zhang, Yuquan Xie, Zaijing Li, Meng Liu, and Liqiang Nie.
\newblock {{ObjectNLQ}}@ {{Ego4D}} episodic memory challenge 2024.
\newblock \emph{arXiv preprint arXiv:2406.15778}, 2024.

\bibitem[Feng et~al.(2025)Feng, Zhang, Liu, Guan, and Nie]{feng2025object}
Yisen Feng, Haoyu Zhang, Meng Liu, Weili Guan, and Liqiang Nie.
\newblock Object-shot enhanced grounding network for egocentric video.
\newblock \emph{arXiv preprint arXiv:2505.04270}, 2025.

\bibitem[Grauman et~al.(2022)Grauman, Westbury, Byrne, Chavis, Furnari, Girdhar, Hamburger, Jiang, Liu, Liu, et~al.]{grauman2022ego4d}
Kristen Grauman, Andrew Westbury, Eugene Byrne, Zachary Chavis, Antonino Furnari, Rohit Girdhar, Jackson Hamburger, Hao Jiang, Miao Liu, Xingyu Liu, et~al.
\newblock Ego4d: {{Around}} the world in 3,000 hours of egocentric video.
\newblock In \emph{Proceedings of the {{IEEE}}/{{CVF Conference}} on {{Computer Vision}} and {{Pattern Recognition}}}, pages 18995--19012, 2022.

\bibitem[Guan et~al.(2022)Guan, Song, Zhang, Liu, Yeh, and Chang]{guan2022bi}
Weili Guan, Xuemeng Song, Haoyu Zhang, Meng Liu, Chung-Hsing Yeh, and Xiaojun Chang.
\newblock Bi-directional heterogeneous graph hashing towards efficient outfit recommendation.
\newblock In \emph{Proceedings of the 30th ACM international conference on multimedia}, pages 268--276, 2022.

\bibitem[Hou et~al.(2023)Hou, Ji, Gao, Zhong, Yan, Li, Chan, Ngo, Duan, and Shou]{hou2023groundnlq}
Zhijian Hou, Lei Ji, Difei Gao, Wanjun Zhong, Kun Yan, Chao Li, Wing-Kwong Chan, Chong-Wah Ngo, Nan Duan, and Mike~Zheng Shou.
\newblock Groundnlq@ ego4d natural language queries challenge 2023.
\newblock \emph{arXiv preprint arXiv:2306.15255}, pages 1--5, 2023.

\bibitem[Liu et~al.(2018{\natexlab{a}})Liu, Wang, Nie, He, Chen, and Chua]{liu2018attentive}
Meng Liu, Xiang Wang, Liqiang Nie, Xiangnan He, Baoquan Chen, and Tat-Seng Chua.
\newblock Attentive moment retrieval in videos.
\newblock In \emph{The 41st international ACM SIGIR conference on research \& development in information retrieval}, pages 15--24, 2018{\natexlab{a}}.

\bibitem[Liu et~al.(2018{\natexlab{b}})Liu, Wang, Nie, Tian, Chen, and Chua]{liu2018cross}
Meng Liu, Xiang Wang, Liqiang Nie, Qi Tian, Baoquan Chen, and Tat-Seng Chua.
\newblock Cross-modal moment localization in videos.
\newblock In \emph{Proceedings of the 26th ACM international conference on Multimedia}, pages 843--851, 2018{\natexlab{b}}.

\bibitem[Liu et~al.(2024)Liu, Sui, Zhang, Mu, Zhao, and Ghanem]{liu2024harnessing}
Shuming Liu, Lin Sui, Chen-Lin Zhang, Fangzhou Mu, Chen Zhao, and Bernard Ghanem.
\newblock Harnessing temporal causality for advanced temporal action detection.
\newblock \emph{arXiv preprint arXiv:2407.17792}, 2024.

\bibitem[Pei et~al.(2024)Pei, Chen, Xu, He, Liu, Pan, Huang, Wang, Lu, Wang, et~al.]{pei2024egovideo}
Baoqi Pei, Guo Chen, Jilan Xu, Yuping He, Yicheng Liu, Kanghua Pan, Yifei Huang, Yali Wang, Tong Lu, Limin Wang, et~al.
\newblock {{EgoVideo}}: {{Exploring}} egocentric foundation model and downstream adaptation.
\newblock \emph{arXiv preprint arXiv:2406.18070}, pages 1--8, 2024.

\bibitem[Plou et~al.(2024)Plou, {Mur-Labadia}, {Martinez-Cantin}, and Murillo]{plou2024carlor}
Carlos Plou, Lorenzo {Mur-Labadia}, Ruben {Martinez-Cantin}, and Ana~C Murillo.
\newblock {{CARLOR}}@ {{Ego4D}} step grounding challenge: {{Bayesian}} temporal-order priors for test time refinement.
\newblock \emph{arXiv preprint arXiv:2406.09575}, pages 1--4, 2024.

\bibitem[Ramakrishnan et~al.(2023)Ramakrishnan, {Al-Halah}, and Grauman]{ramakrishnan2023naq}
Santhosh~Kumar Ramakrishnan, Ziad {Al-Halah}, and Kristen Grauman.
\newblock {{NaQ}}: {{Leveraging}} narrations as queries to supervise episodic memory.
\newblock In \emph{Proceedings of the {{IEEE}}/{{CVF Conference}} on {{Computer Vision}} and {{Pattern Recognition}}}, pages 6694--6703. IEEE Computer Society, 2023.

\bibitem[Song et~al.(2024)Song, Byrne, Nagarajan, Wang, Martin, and Torresani]{song2024ego4d}
Yale Song, Eugene Byrne, Tushar Nagarajan, Huiyu Wang, Miguel Martin, and Lorenzo Torresani.
\newblock Ego4d goal-step: {{Toward}} hierarchical understanding of procedural activities.
\newblock \emph{Advances in Neural Information Processing Systems}, 36, 2024.

\bibitem[Wang et~al.(2025)Wang, Liu, Shao, Zhang, Wen, Yang, Gao, Zhang, and Nie]{wang2025time}
Yunxiao Wang, Meng Liu, Rui Shao, Haoyu Zhang, Bin Wen, Fan Yang, Tingting Gao, Di Zhang, and Liqiang Nie.
\newblock Time: Temporal-sensitive multi-dimensional instruction tuning and benchmarking for video-llms.
\newblock \emph{arXiv preprint arXiv:2503.09994}, 2025.

\bibitem[Zeng et~al.(2025)Zeng, Zhong, Feng, and Ma]{zeng2025unimd}
Yingsen Zeng, Yujie Zhong, Chengjian Feng, and Lin Ma.
\newblock Unimd: {{Towards}} unifying moment retrieval and temporal action detection.
\newblock In \emph{European Conference on Computer Vision}, pages 286--304. Springer, 2025.

\bibitem[Zhang et~al.(2025{\natexlab{a}})Zhang, Sui, Liu, Mu, Wang, and Ghanem]{zhang2025timeloc}
Chen-Lin Zhang, Lin Sui, Shuming Liu, Fangzhou Mu, Zhangcheng Wang, and Bernard Ghanem.
\newblock {{TimeLoc}}: A unified end-to-end framework for precise timestamp localization in long videos.
\newblock \emph{arXiv preprint arXiv:2503.06526}, 2025{\natexlab{a}}.

\bibitem[Zhang et~al.(2021{\natexlab{a}})Zhang, Liu, Gao, Lei, Wang, and Nie]{10.1145/3474085.3475234}
Haoyu Zhang, Meng Liu, Zan Gao, Xiaoqiang Lei, Yinglong Wang, and Liqiang Nie.
\newblock Multimodal dialog system: Relational graph-based context-aware question understanding.
\newblock In \emph{Proceedings of the 29th ACM International Conference on Multimedia}, page 695–703. Association for Computing Machinery, 2021{\natexlab{a}}.

\bibitem[Zhang et~al.(2021{\natexlab{b}})Zhang, Liu, Gao, Lei, Wang, and Nie]{zhang2021multimodal}
Haoyu Zhang, Meng Liu, Zan Gao, Xiaoqiang Lei, Yinglong Wang, and Liqiang Nie.
\newblock Multimodal dialog system: Relational graph-based context-aware question understanding.
\newblock In \emph{Proceedings of the 29th ACM international conference on multimedia}, pages 695--703, 2021{\natexlab{b}}.

\bibitem[Zhang et~al.(2023{\natexlab{a}})Zhang, Liu, Li, Yan, Gao, Chang, and Nie]{10239469}
Haoyu Zhang, Meng Liu, Yuhong Li, Ming Yan, Zan Gao, Xiaojun Chang, and Liqiang Nie.
\newblock Attribute-guided collaborative learning for partial person re-identification.
\newblock \emph{IEEE Transactions on Pattern Analysis and Machine Intelligence}, 45\penalty0 (12):\penalty0 14144--14160, 2023{\natexlab{a}}.

\bibitem[Zhang et~al.(2023{\natexlab{b}})Zhang, Liu, Wang, Cao, Guan, and Nie]{zhang2023uncovering}
Haoyu Zhang, Meng Liu, Yaowei Wang, Da Cao, Weili Guan, and Liqiang Nie.
\newblock Uncovering hidden connections: Iterative tracking and reasoning for video-grounded dialog.
\newblock \emph{arXiv preprint arXiv:2310.07259}, 2023{\natexlab{b}}.

\bibitem[Zhang et~al.(2024{\natexlab{a}})Zhang, Liu, Liu, Song, Wang, and Nie]{pmlr-v235-zhang24aj}
Haoyu Zhang, Meng Liu, Zixin Liu, Xuemeng Song, Yaowei Wang, and Liqiang Nie.
\newblock Multi-factor adaptive vision selection for egocentric video question answering.
\newblock In \emph{Proceedings of the 41st International Conference on Machine Learning}, pages 59310--59328. PMLR, 2024{\natexlab{a}}.

\bibitem[Zhang et~al.(2024{\natexlab{b}})Zhang, Xie, Feng, Li, Liu, and Nie]{zhang2024hcqa}
Haoyu Zhang, Yuquan Xie, Yisen Feng, Zaijing Li, Meng Liu, and Liqiang Nie.
\newblock Hcqa@ ego4d egoschema challenge 2024.
\newblock \emph{arXiv preprint arXiv:2406.15771}, 2024{\natexlab{b}}.

\bibitem[Zhang et~al.(2025{\natexlab{b}})Zhang, Chu, Liu, Wang, Wen, Yang, Gao, Zhang, Wang, and Nie]{zhang2025exo2ego}
Haoyu Zhang, Qiaohui Chu, Meng Liu, Yunxiao Wang, Bin Wen, Fan Yang, Tingting Gao, Di Zhang, Yaowei Wang, and Liqiang Nie.
\newblock Exo2ego: Exocentric knowledge guided mllm for egocentric video understanding.
\newblock \emph{arXiv preprint arXiv:2503.09143}, 2025{\natexlab{b}}.

\bibitem[Zhang et~al.(2025{\natexlab{c}})Zhang, Feng, Chu, Liu, Guan, Wang, and Nie]{zhang2025hcqa}
Haoyu Zhang, Yisen Feng, Qiaohui Chu, Meng Liu, Weili Guan, Yaowei Wang, and Liqiang Nie.
\newblock Hcqa-1.5@ ego4d egoschema challenge 2025.
\newblock \emph{arXiv preprint arXiv:2505.20644}, 2025{\natexlab{c}}.

\end{thebibliography}
}


\end{document}